\documentclass[10pt,twocolumn,letterpaper]{article}

\usepackage{wacv}
\usepackage{times}
\usepackage{epsfig}
\usepackage{graphicx} 
\usepackage{amsmath}
\usepackage{amssymb}
\usepackage{subcaption}
\usepackage{caption}
\usepackage{booktabs,longtable,tabu}
\usepackage{csquotes} 
\usepackage{wrapfig,lipsum}
\usepackage[pagebackref=true,breaklinks=true,letterpaper=true,colorlinks,bookmarks=false]{hyperref}

\wacvfinalcopy % *** Uncomment this line for the final submission

\def\wacvPaperID{203} % *** Enter the wacv Paper ID here

\ifwacvfinal\fi
\begin{document}

%%%%%%%%% TITLE
\title{Unsupervised Adversarial Visual Level Domain Adaptation for Learning Video Object Detectors from Images }

% Authors at the same institution
%\author{First Author \hspace{2cm} Second Author \\
% Authors at different institutions
\author{Avisek Lahiri* \\
Dept. of E\&ECE, IIT Kharagpur\\
{\tt\small avisek@ece.iitkgp.ernet.in}
\and
Charan Reddy* \\
Dept. of CSE, IIT Kharagpur\\
{\tt\small ragireddycharan223@gmail.com }
\and
Prabir Kumar Biswas \\
Dept. of E\&ECE, IIT Kharagpur\\
{\tt\small pkb@ece.iitkgp.ernet.in}
\thanks{Equal contribution}
}

\maketitle
\ifwacvfinal\thispagestyle{empty}\fi

%%%%%%%%% ABSTRACT
\begin{abstract}
   Deep learning based object detectors require thousands of diversified bounding box and class annotated examples. Though image object detectors have shown rapid progress in recent years with release of multiple large scale static image datasets, object detection on videos still remains an open problem due to scarcity of annotated video frames. Having a robust video object detector is an essential component for video understanding and curating large scale automated annotations in videos. Domain difference between images and videos makes the transferability of image object detectors to videos sub-optimal. The most common solution is to use weakly supervised annotations where a video frame has to be tagged for presence/absence of object categories. This still takes up manual effort. In this paper we take a step forward by adapting the concept of unsupervised adversarial image-to-image translation to perturb static high quality images to be visually indistinguishable from a set of video frames. We assume the presence of a fully annotated static image dataset and an unannotated video dataset. Object detector is trained on adversarially transformed image dataset using the annotations of the original dataset. Experiments on Youtube-Objects and Youtube-Objects-Subset datasets with two contemporary baseline object detectors reveal that such unsupervised pixel level domain adaptation boosts the generalization performance on video frames compared to direct application of original image object detector. Also, we achieve competitive performance compared to recent baselines of weakly supervised methods. This paper can be seen as an application of image translation for cross domain object detection.
\end{abstract}
%===========================
\begin{figure}[!t]
    \centering
    \includegraphics[scale=0.2]{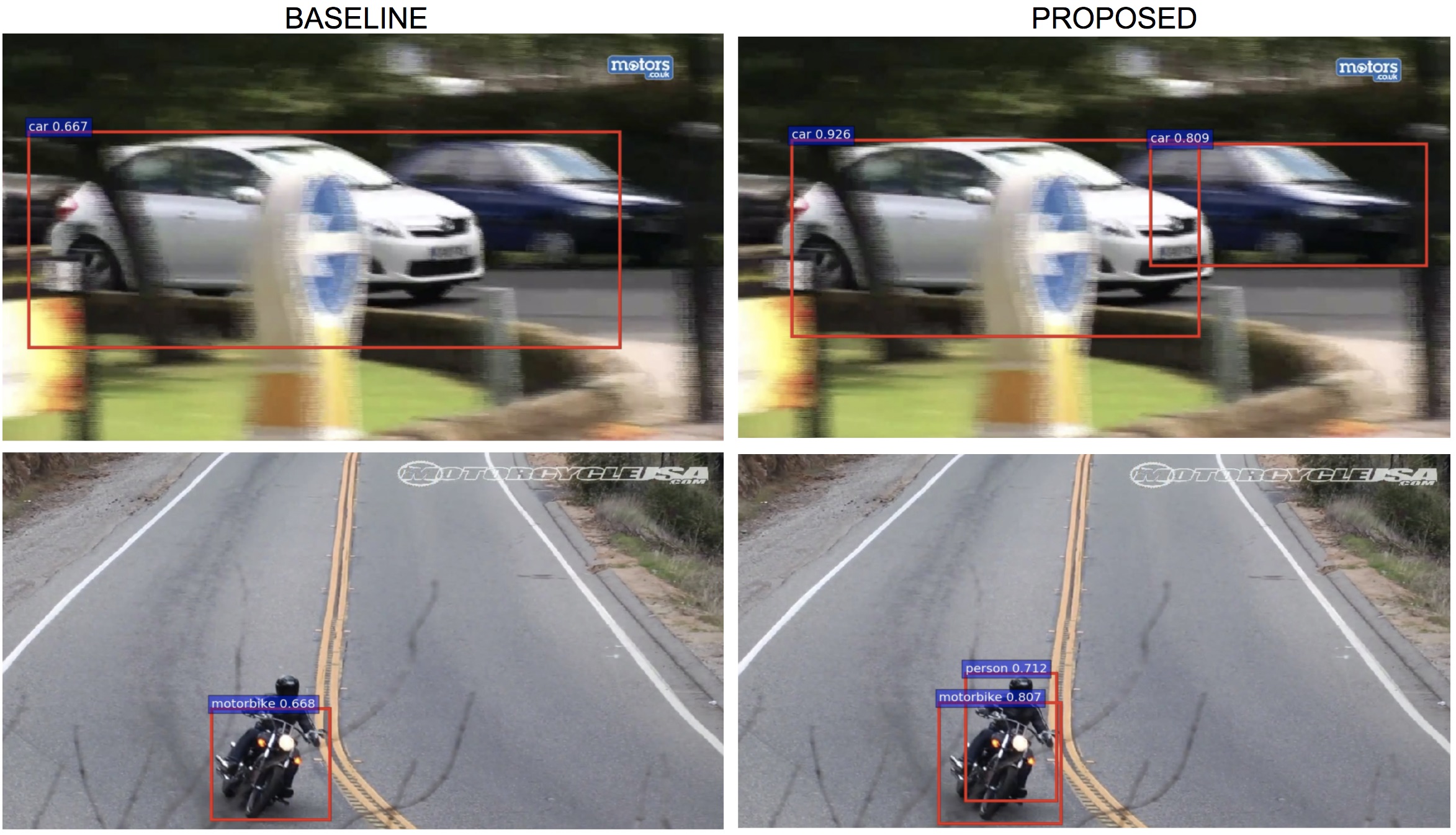}  
    \caption{Exemplary success of our model in detecting objects in high speed and blurry video frames. \textbf{Left:} Detection by Faster R-CNN \cite{faster} object detector trained on static images of PASCAL VOC. \textbf{Right:} Detection by the same Faster R-CNN framework but trained on our proposed adversarially transformed PASCAL VOC images. Our video class agnostic framework achieves near to state-of-the-art results compared to weakly supervised methods.}%
    \label{fig_cover}%
\end{figure}
%====================================
%===========================
\begin{figure*}[!t]
    \centering
    \includegraphics[scale=0.53]{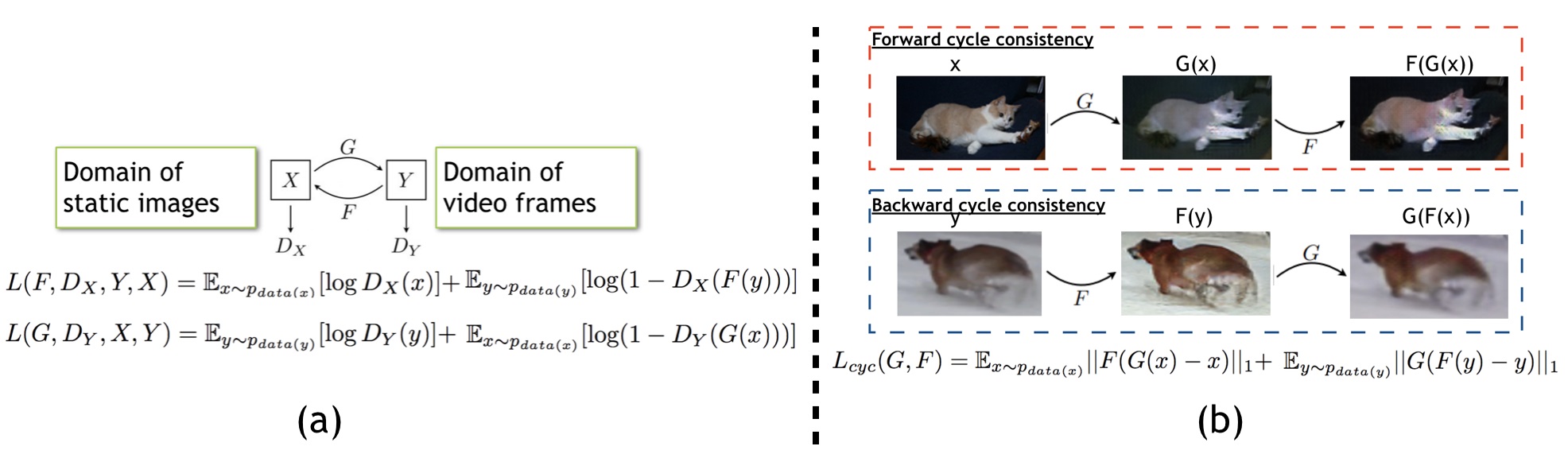}  
    \caption{(a) Our model consists of two transformation functions, $G(\cdot)$ and $F(\cdot)$ to transform images sampled from domain (X) of high quality images to be visually indistinguishable from samples of the domain(Y) of video frames and vice versa. Discriminator $D_Y$, discriminates between $G(X)$ and $Y$ while $D_X$ discriminates between $X$ and $F(Y)$. (b) To limit the set of possible outputs for the under constrained mappings, $G(\cdot)$ and $F(\cdot)$, two cycle consistency losses are enforced such that $x \rightarrow G(x) \rightarrow F(G(x)) \approx x$ and $y \rightarrow F(y) \rightarrow G(F(y)) \approx y$.}%
    \label{fig_flow}%
\end{figure*}
%====================================
%%%%%%%%% BODY TEXT
\section{Introduction}
We consider the problem of object detection in unconstrained videos with close to zero supervision for video domain. Object detection in videos is a crucial component in several downstream vision applications such as video anomaly detection, autonomous driving, tracking etc. In this work, we assume that we only have access to a fully annotated dataset of still images (which we refer to as \textit{source domain}) while there is no annotation (not even any form of weak supervision) for video domain (which we refer to as \textit{target domain}). Intuitively, an object detector trained on still images performs worse on video frames primarily due to significant appearance disparities between the two domains. Specifically, still images retain much more high frequency components and are less cluttered/occluded. Conversely, objects in videos often suffer from motion blur and poor resolution. Thus even though we have availability of large-scale annotated image datasets such as ImageNet\cite{deng2009imagenet}, MS-COCO\cite{lin2014microsoft}, PASCAL \cite{everingham2010pascal}, performance of image object detectors on videos are worse compared to training on manually annotated video datasets \cite{gokberk2014multi, song2014learning}. An immediate direction of effort can be to annotate video datasets. However, annotating large scale video datasets demand humongous manual labor, time and cost.
\par Though collecting annotated video seems a daunting task, there is abundance of unlabeled natural videos available publicly from sources such as Youtube and Flickr. The aim of this paper is to exploit such unlabeled videos to learn an end-to-end trainable network to transform images sampled from static image dataset to `appear' as if being sampled from a video dataset. Following this, if we train an object detector on such `\textit{transformed image dataset}', we expect to see a boost in the generalization capability of the detector on videos(See Fig. \ref{fig_cover} for exemplary success). Specifically our contributions in this paper can be summarized as:
\begin{enumerate}
\item We apply the concept of cycle consistent image-to-image translation\cite{zhu2017unpaired} with generative adversarial networks(GAN) \cite{goodfellow2014generative} for learning a completely unsupervised transformation from image to video in pixel domain. To our best knowledge, this work is the first demonstration of the applicability of GAN based pixel level domain adaptation for adapting object detectors across image to video.
\item We empirically show the importance of cyclic network architecture for training an unsupervised GAN based image translation framework. 
\item Evaluations on recently released Youtube-Objects \cite{kalogeiton2016analysing} and Youtube-Objects-Subset \cite{tang2013discriminative} datasets reveal that our approach of domain adaptation improves upon two contemporary baseline state-of-the-art image object detectors. Also, we get competitive performance compared to recent weakly supervised methods.
\end{enumerate}
%===================================
%=================================
\begin{figure*}[!t]
    \centering
    \includegraphics[scale=0.3]{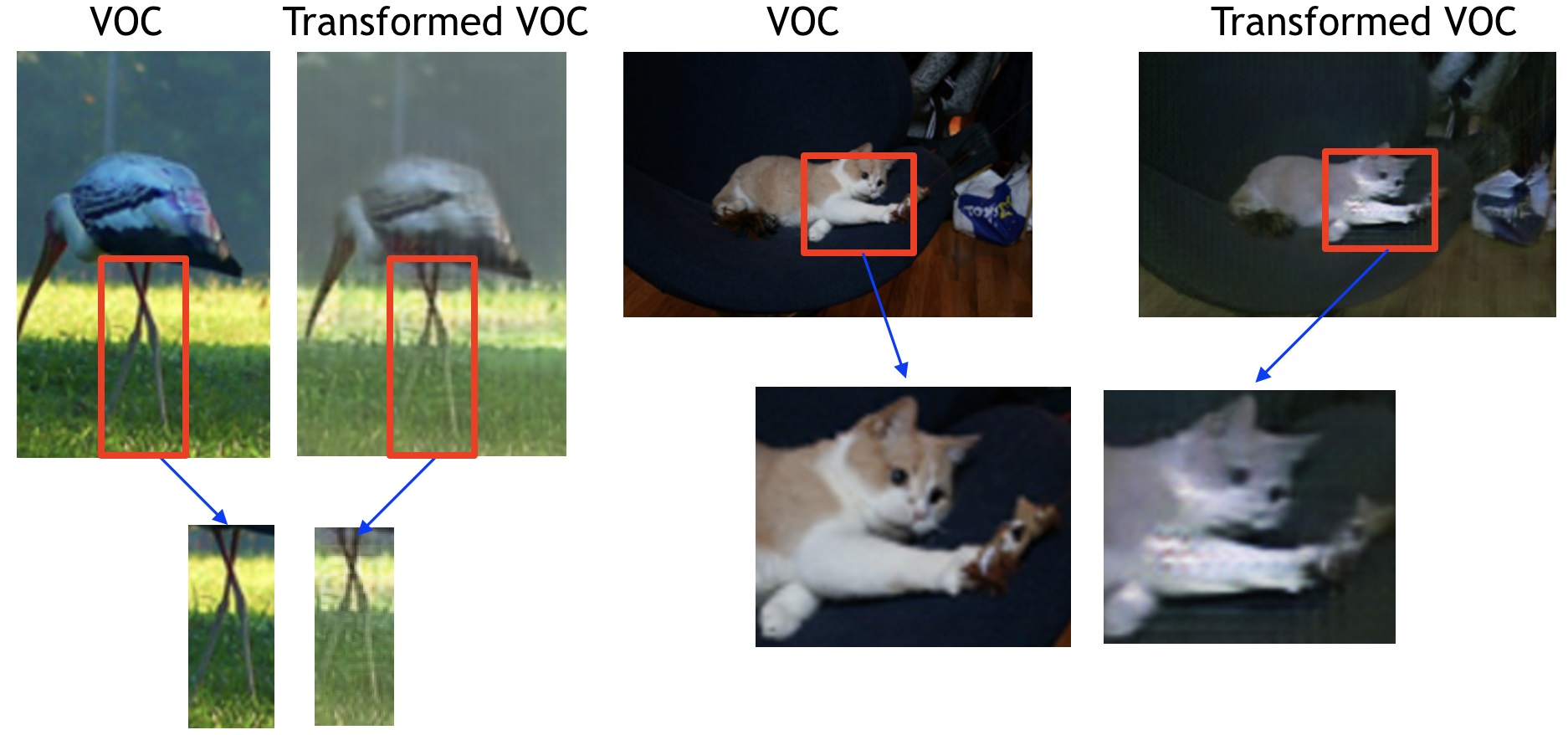} 
    \caption{Effect of applying learnt transformations on high quality static VOC images to appear as if being sampled from a video sequence. Note (left) how the legs of the bird get almost blended with nearby surroundings after transformation. Also, we show a case(right) where discriminative fur colors of a cat get desaturated and details such as eyes, whiskers and ears becomes indistinguishable. Training object detectors on such adversarially perturbed images helps in improving test performance on actual video frames. Best appreciated when viewed in color and zoomed up.}%
    \label{fig_effect}% 
\end{figure*}
%====================================
\section{Related Works}
\subsection{Object Detection}
Object detection in still images is one of the traditional genres of computer vision  research \cite{dalal2005histograms, felzenszwalb2010object,girshick2014rich, malisiewicz2011ensemble}. These methods require bounding box annotations on a large sample of diversified images. Although annotating boxes for large datasets is tedious it is necessary especially for deep learning frameworks in which the complexity of the neural net makes it vulnerable to overfitting if not trained with sufficiently large labeled datasets. To circumnavigate this requirement there are two genres of approach closely related to our current effort. One line of approach is weakly supervised object localization \cite{bilen2014weakly,kumar2010self, prest2012learning,rochan2015weakly,vcip,bmvc}, wherein we only have meta information such as the presence/absence of an object category. Majority of these algorithms are based on multiple instance learning(MIL) framework. In this formulation, an image is represented as a bag of regions. It is assumed if an object category is marked as `present' in an image then one of the regions in positive bag tightly bound the object while a tag of `absent' mean no regions contain the object. MIL training alternates between learning an appearance model and selecting the proper region in positive bags which contain the object using the learnt appearance model. However, weakly supervised training is very difficult and its performance is still not at par with fully supervised methods \cite{gokberk2014multi, song2014learning}. 
\par The second line of effort is to exploit information from both videos and images \cite{prest2012learning, sharma2013efficient, leistner2011improving}. In \cite{prest2012learning}, Prest \textit{et al.} presented a framework for adapting detectors from videos to images (which is opposite to ours). First, they learn an automated video object localizer in videos. This is followed by training video object detector under a domain adaptation setting with fully annotated image data and weakly annotated video data. In \cite{sharma2013efficient}, Sharma and Nevatia present a framework for detecting humans in videos by online refinement of a baseline pre-trained detector. The idea is to apply the pre-trained detector at a high recall rate so that it outputs a lot of true positive regions. The false positive regions are discarded by an online refinement. Clearly, this method is not suited for real-time applications. Our detector modules consist of Faster R-CNN \cite{faster} which runs in near real time. Also, we experiment on wide varieties of moving objects(rather than only humans) under more unconstrained environments.
%=======================
\subsection{Generative Adversarial Networks}
Generative adversarial network\cite{goodfellow2014generative} has two parametrized models, discriminator and generator pitted against each other in a zero-sum game. The generator network's input is a noise vector, $z$, drawn from a prior noise distribution, $p_z(z)$. Following \cite{goodfellow2014generative},  $z\sim \mathcal{U}[-1,1]$ (uniform distribution) and generator maps it onto an image, $y$; $G~: z~ \rightarrow y$. The discriminator classifies samples between true data distribution $p_{data}$ and the generated distribution, $p_G$. Specifically, generator and discriminator play the following game on $V(D, G)$:
$$\underset{G}{min}~~ \underset{D}{max}~~ V(D, G) = \mathbb{E}_{x\sim p_{data}(x)}[\log D(x)]$$
\begin{equation}
+ ~~\mathbb{E}_{z\sim p_{z}(z)}[1 - D(G(z))]
\end{equation}
With sufficient capacity for both generator and discriminator, this min-max game has global optimum when $p_{data}=p_G$ \cite{goodfellow2014generative}. Empirically, it has been observed that for the generator, it is prudent to maximize $\log (D(G(z)))$ instead of minimizing $\log [1 - D(G(z))] $.
%=========================
\subsection{Image to Image Translation with Adversarial Learning}
Our core motivation is to transform a static image to appear as if being sampled from a dataset of video frames. This can be seen as an appearance mapping in pixel space, which is nowadays studied under the umbrella of adversarial image-to-image translation genre. GANs show the potential to approximate an unknown static distribution. This property has been recently leveraged by several authors for pixel level domain adaptation \cite{shrivastava2017learning, pix2pix2017, bousmalis2017unsupervised}. The basic idea is to have paired image samples in both domains, $X$ and $Y$ and then learn a conditional generator network(conditioned on an image sampled from $X$) to map to $Y$. The discriminator's task is to identify whether an image is sampled from $Y$ or is transformed from $X$. Shrivastava \textit{et al.} \cite{shrivastava2017learning} and Bousmalis \textit{et al.} \cite{bousmalis2017unsupervised} had the common motivation to design a `refiner' network to transform synthetic images to appear like real images and then train discriminative models on these transformed synthetic datasets. This is helpful because getting labelled data in a rendered/synthetic domain is often free of cost. Though promising, these methods were only applied for inferencing on very small objects such as estimating gaze from a properly cropped out human eye sample of resolution 35$\times$55. In \cite{bousmalis2017unsupervised}, the authors presented results of recognition and pose estimation from specifically centre cropped small objects such as `phone', `lamp' etc., from Cropped LineMod dataset \cite{wohlhart2015learning}. Our application case is much more difficult as we are working with non cropped image/video frames in the wild with the problem of detecting an arbitrary number of instances of a given class in an image. The first success of applying GANs for high resolution (256$\times$ 256) image-to-image translation was proposed by Isola \textit{et al.} \cite{pix2pix2017}. Their framework, for example, transforms sketch of a shoe to real textured consumer shoe or converts an aerial map to actual city image.  Though promising, this method is restricted in applicability due to the requirement of paired examples across both domains. This is specifically restrictive in our case because it is not possible to have an object with the same scale and orientation to be present in both image and video datasets. To mitigate this restriction, Zhu \textit{et al.} \cite{zhu2017unpaired} proposed to incorporate a cycle consistency loss so that a forward transform, $F(X)$, followed by a backward transform, G(F(x)), gives back the starting distribution, $X$. The same restriction is applied for domain $Y$. The cycle consistency loss is a key component to learn in absence of paired data and is thus well suited for our application use case. 
%=============================================================================
\section{Our Approach}
Our aim is to learn an unsupervised mapping between two domains of data, viz., high-quality static images, $X$, and the other domain of video frames, $Y$. We have unpaired training samples, $x_1, x_2,... , x_N \in X$ and $y_1, y_2,... , y_M \in Y$. We denote data distribution as $x \sim p_{data}(x)$ and $y \sim p_{data}(y)$. In Fig. \ref{fig_flow} we show the two components of adversarial transformation and cycle consistency loss between image and video domains. There are two transformation networks, $G(\cdot)$ and $F(\cdot)$, which map  $X \rightarrow Y$ and $Y \rightarrow X$ respectively. 
%==================== 
\subsection{Adversarial Transformation}
Domain discriminator, $D_Y$, discriminates between images transformed from the static images and frames sampled from videos. The adversarial loss for this forward transformation is defined as,
$$L(G, D_Y, X, Y) = \mathbb{E}_{y \sim p_{data(y)}} [\log D_Y(y)] +$$
\begin{equation}
\mathbb{E}_{x \sim p_{data(x)}} [\log (1 - D_Y(G(x)))]
\label{eq_d_y}
\end{equation}
$D_X$ discriminates between video frames transformed to image domain and images sampled from static image database. The corresponding adversarial loss is,
$$L(F, D_X, Y, X) = \mathbb{E}_{x \sim p_{data(x)}} [\log D_X(x)] + $$
\begin{equation}
\mathbb{E}_{y \sim p_{data(y)}} [\log (1 - D_X(F(y)))]
\label{eq_d_x}
\end{equation}
%======================
\subsection{Cycle Consistency Loss}
Theoretically, with enough capacity, $G$ and $F$ can learn to generate samples from $Y$ and $X$ respectively. However, without any additional constraint, a given sample from $X$ can be mapped to any random point in $Y$ and be indistinguishable from real samples of $Y$. For example, if we provide an image of a car from the static dataset, $G$ can map it to look like a car from video dataset, but can change the scale and pose of the car. Though, this might not be an issue from an artistic point of view, it is a point of concern for training object detectors because we will use the bounding box annotations from the  original static image. Thus we cannot afford to have any structural change during mapping from $X \rightarrow Y$. To restrict the domain of possible transformations, a cycle loss is introduced such that learned mapping respects the following sequence of transformation constraint.
\begin{equation}
x \rightarrow G(x) \rightarrow F(G(x)) \approx x.
\end{equation}
This ensures that the learned cyclic mapping can start from a given image $x\sim X$ and we get back $x$ after the two transformations of the cycle. This is termed as the forward cycle consistency constraint and the corresponding loss is,
\begin{equation}
L_{fwd}(G, F) = \mathbb{E}_{x \sim p_{data(x)}}||F(G(x)-x)||_1
\end{equation}
A similar consistency is also imposed for frames being converted to static images with $F(\cdot)$ and back to video frames with $G(\cdot)$. The following constraint needs to be maintained,
\begin{equation}
y \rightarrow F(y) \rightarrow G(F(y)) \approx y
\end{equation}
and the corresponding backward consistency loss is 
\begin{equation}
L_{bwd}(G,F) = \mathbb{E}_{y \sim p_{data(y)}}||G(F(y)-y)||_1
\end{equation}
%========================================
\subsection{Complete Objective}
The complete objective function can be written as,
$$L(G, F, D_X, D_Y) = L(G, D_Y, X, Y) + L(F, D_X, Y, X) + $$
\begin{equation}
\lambda(L_{fwd}(G, F) + L_{bwd}(G,F)),
\end{equation}
where $\lambda$ controls the relative importance of cycle consistency loss over the adversarial loss. The task is to find optimum, $G^*$ and $F^*$ such that,
\begin{equation}
G^*, F^* = \underset{G, F}{\arg\min} \underset{D_X, D_Y}{\max}   L(G, F, D_X, D_Y)
\end{equation} 
%======================================================================
\section{Implementation Details}
Our entire framework consists of two phases of training. In the first phase, we train the CycleGAN network to transform high-quality static images to appear as video frames. So, a given image dataset is transformed into a pseudo video dataset. The next phase is training a standard object detector on the previously transformed image dataset using the same annotations of the original static image dataset. CycleGAN is not required during object detection testing.
%=======================
\subsection{Object Detector}
In this paper we have used two contemporary object detectors viz., Faster R-CNN \cite{faster}\footnote{Available at: \href{https://github.com/smallcorgi/Faster-RCNN_TF}{https:\/\/github.com\/smallcorgi\/Faster-RCNN\_TF}} and RFB Net \cite{liu2017receptive} \footnote{Available at: \href{https://github.com/lzx1413/PytorchSSD}{https:\/\/github.com\/lzx1413\/PytorchSSD}}. We have used the default settings in the respective papers for training on all different variants of the training datasets.
%======================================
%==============================================
\begin{figure*}[!t]
    \centering
    \includegraphics[scale=0.36]{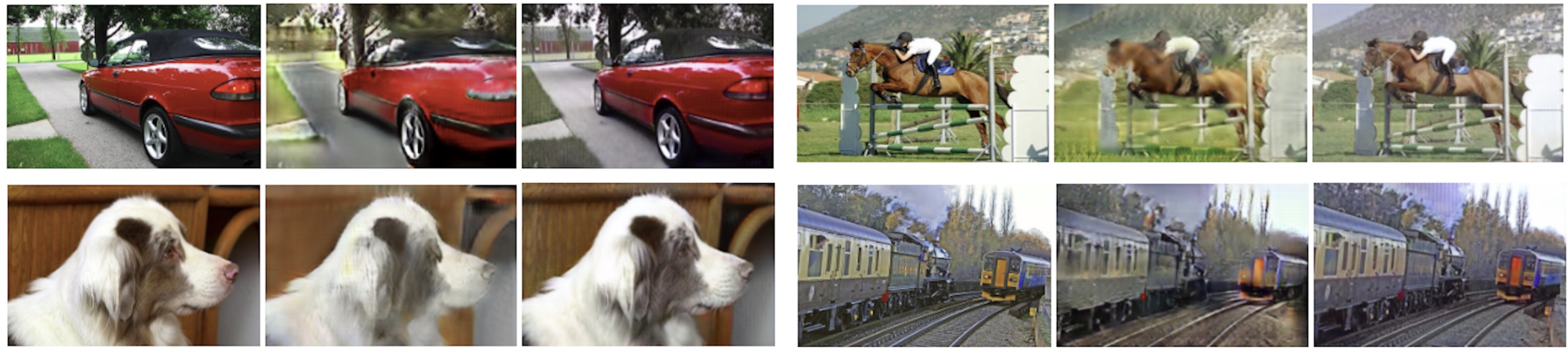}  
    \caption{Benefit of cycle consistent GAN for image to image translation over a simple forward transform. In each tuple, left column: original VOC image, middle column: ForwardGAN transformed image and right column: transformed image by CycleGAN. A ForwardGAN only itself is not able to maintain the structural information in this high dimensional space. The degradations by this framework are not representative of frames from videos and thus training object detectors on these transformed images yield inferior results. Visually, CycleGAN does not degrade the essential structural information of the transformed images but still incorporates necessary perturbations to become indistinguishable from YTO frames. Training object detectors on CycleGAN transformed images thus results in better performance than training on ForwardGAN transformed images. }%
    \label{fig_gans}%
\end{figure*}
%==================================================
\subsection{CycleGAN}
We have trained CycleGAN, on unaligned images and video frames. During training, images are resized to 286$\times$286 and then randomly cropped to 256$\times$256 to increase the robustness of the model. During inference, transformation is done on 256$\times$256 resized image and then rescaled to the original resolution before object detection.
We have modelled the generator with 9 resnet blocks as implemented in \cite{he2016deep} and the discriminator with PatchGAN classifier \cite{pix2pix2017, li2016precomputed}\footnote{Implementation: \href{https://github.com/junyanz/CycleGAN}{https://github.com/junyanz/CycleGAN}}. GAN loss is implemented with vanilla GAN \cite{goodfellow2014generative} objective, while cycle loss is L$_1$ loss. We have considered $\lambda$ = 10. We have considered a batch size of 1 and used Adam optimizer \cite{kingma2014adam} with a momentum of 0.5 and an initial learning rate of 0.0002.
%================================
\subsection{Timings}
All computations are performed on NVIDIA Tesla K40C 12GB GPU.
Training of CycleGAN with 5000 images each in source and target domain takes 2 days. Faster R-CNN training on image dataset takes 12 hours, while RFB Net training takes 10 hours. During testing, on average, Faster R-CNN and RFB Net run at about 7/8 frames per second.
%=======================
\section{Experiments}
In the first part, we show the visual effects of applying adversarial transformation on static images and in second half we visually and numerically show the benefit of pixel level domain adaptation for object detection in videos.
%=================
%===============================================
\begin{figure*}[!h]
\centering
\begin{minipage}[t]{0.38\textwidth}
\includegraphics[width=\linewidth]{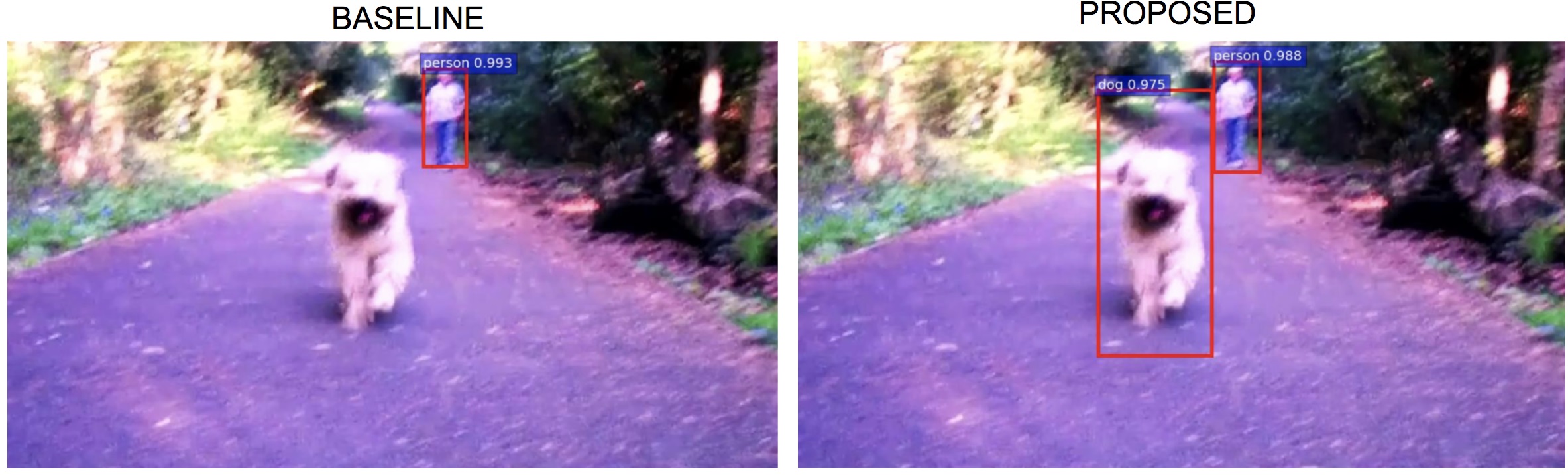}
\end{minipage}
\hspace{2mm}
\begin{minipage}[t]{0.38\textwidth}
\includegraphics[width=\linewidth]{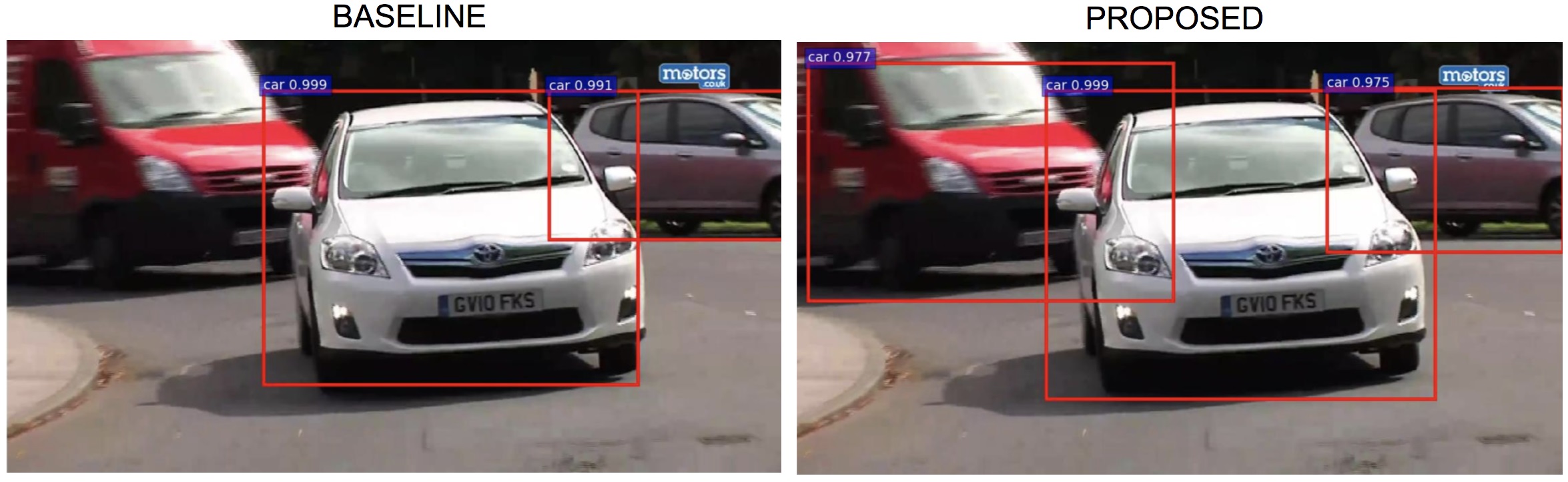}
\end{minipage}
\begin{minipage}[t]{0.38\textwidth}
\includegraphics[width=\linewidth]{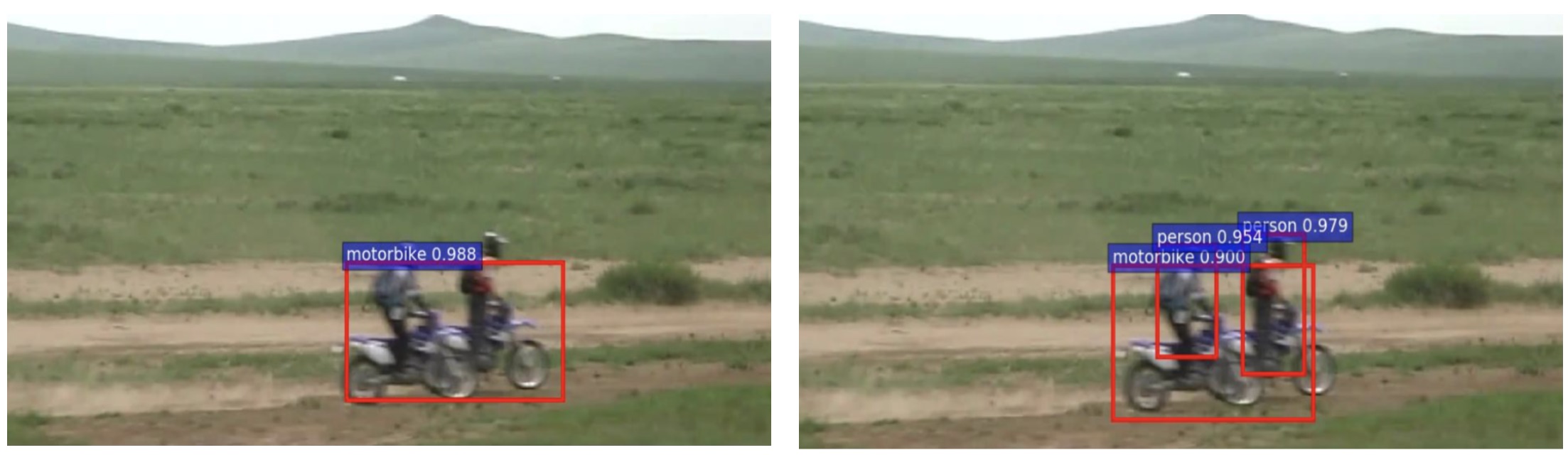}
\end{minipage}
\hspace{2mm}
\begin{minipage}[t]{0.38\textwidth}
\includegraphics[width=\linewidth]{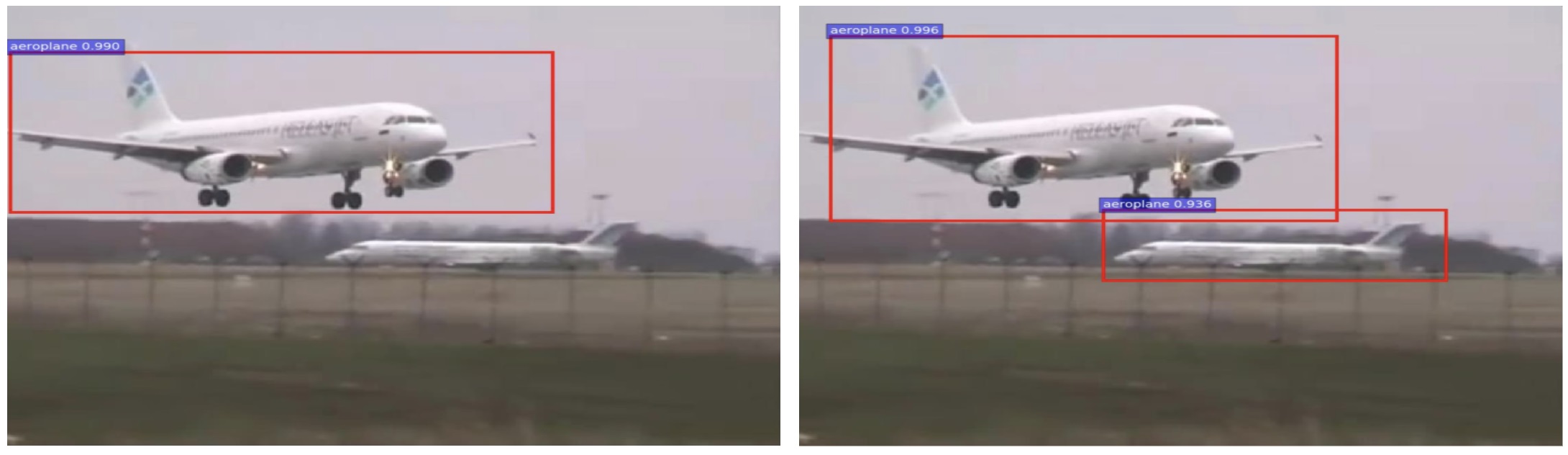}
\end{minipage}
\begin{minipage}[t]{0.38\textwidth}
\includegraphics[width=\linewidth]{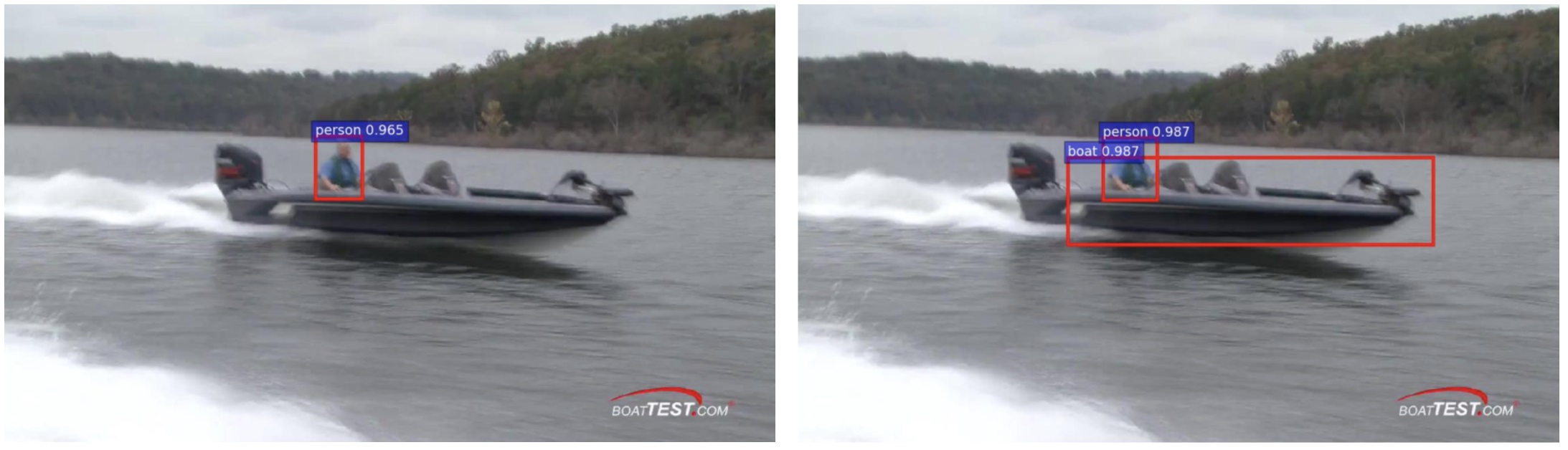}
\end{minipage}
\hspace{2mm}
\begin{minipage}[t]{0.38\textwidth}
\includegraphics[width=\linewidth]{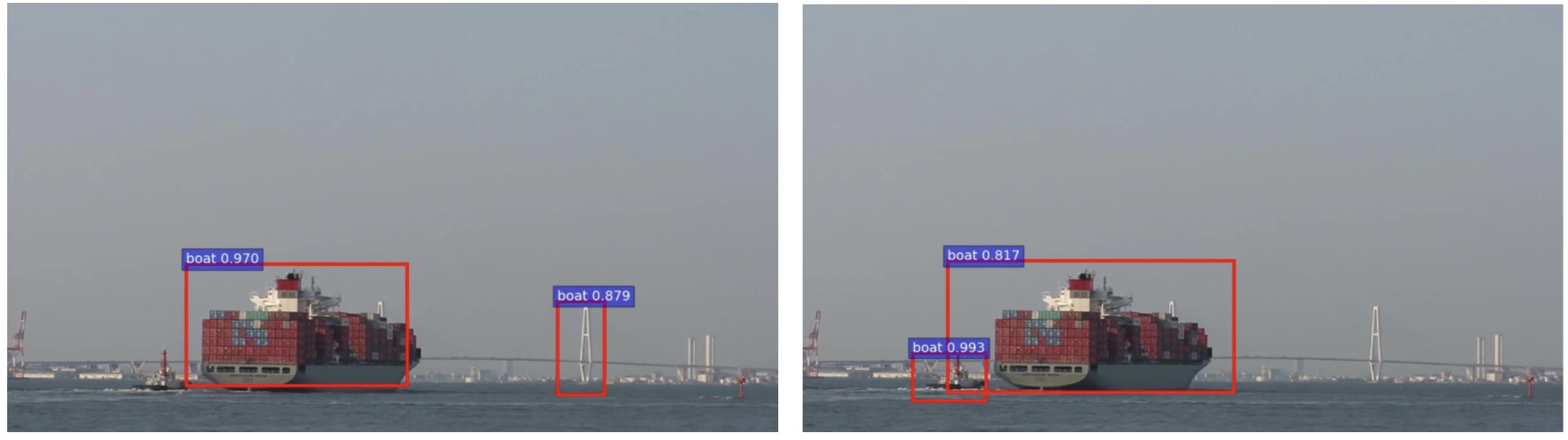}
\end{minipage}
\begin{minipage}[t]{0.38\textwidth}
\includegraphics[width=\linewidth]{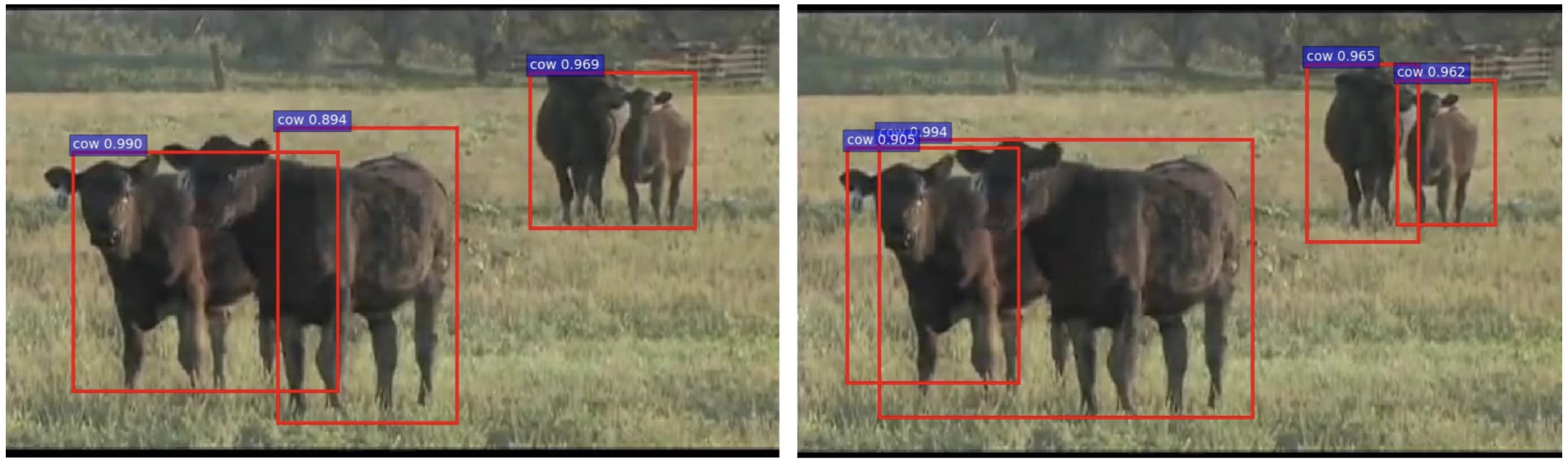}
\end{minipage}
\hspace{2mm}
\begin{minipage}[t]{0.38\textwidth}
\includegraphics[width=\linewidth]{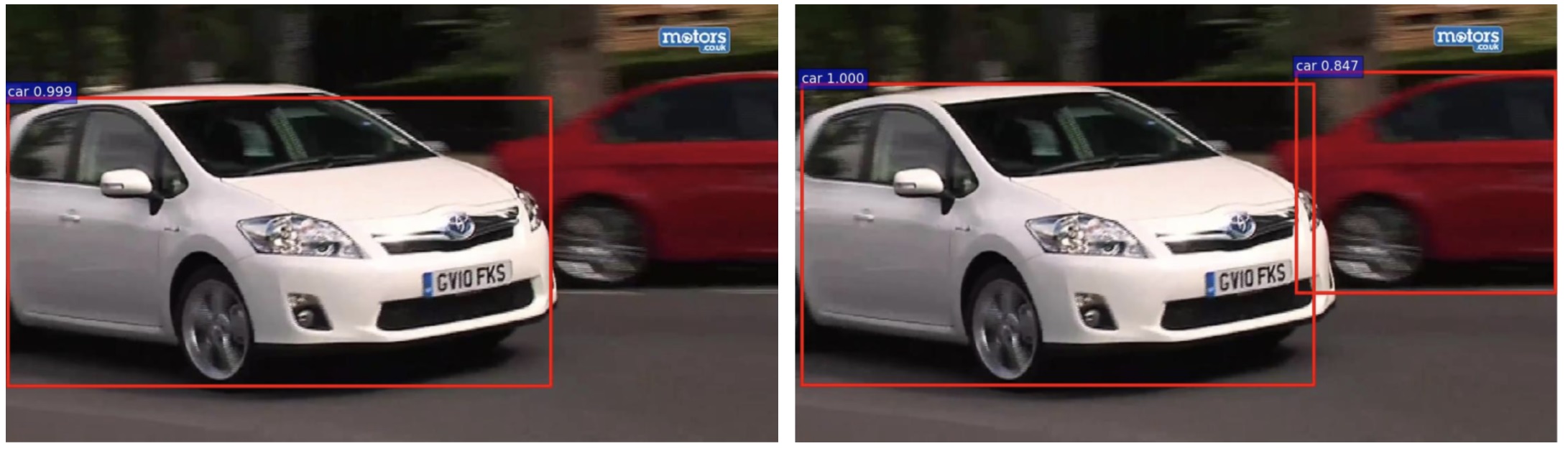}
\end{minipage}
\begin{minipage}[t]{0.38\textwidth}
\includegraphics[width=\linewidth]{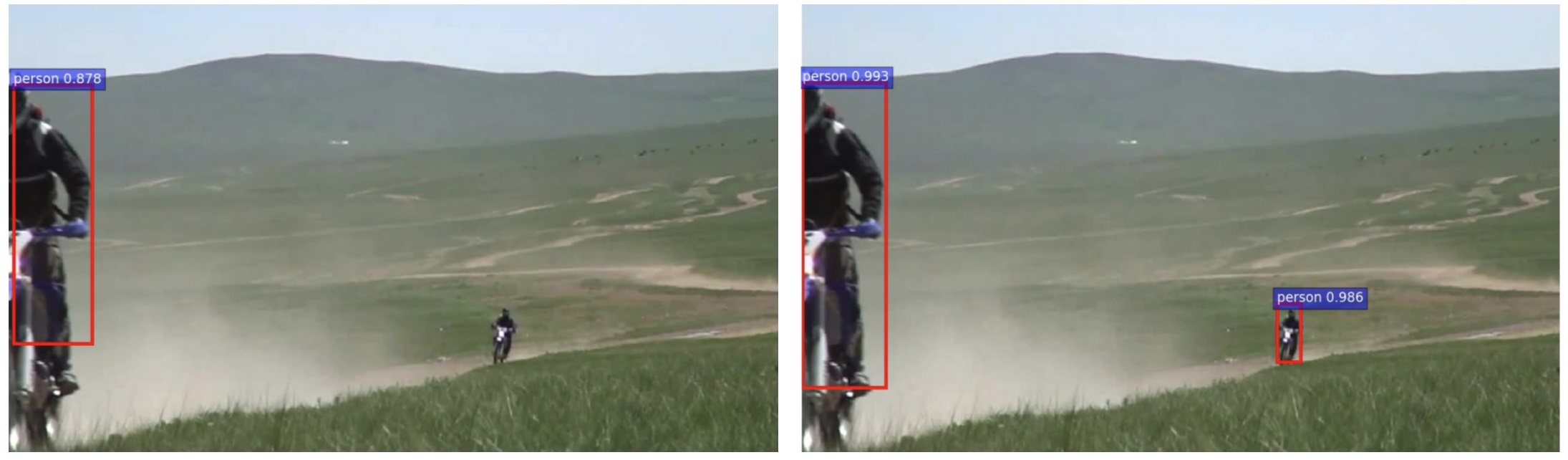}
\end{minipage}
\hspace{2mm}
\begin{minipage}[t]{0.38\textwidth}
\includegraphics[width=\linewidth]{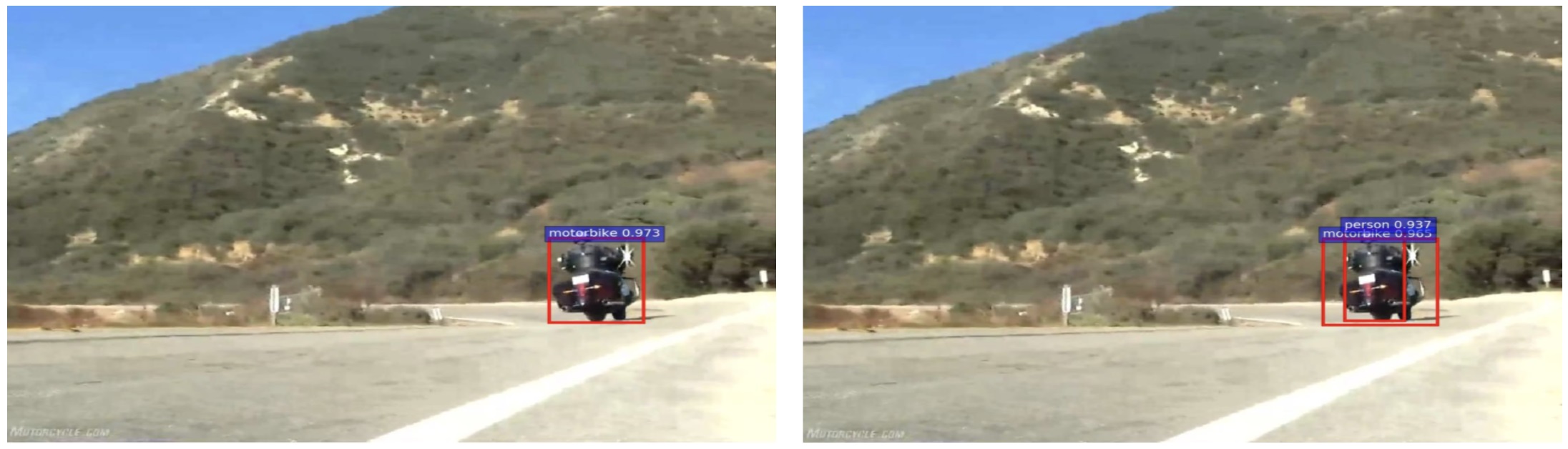}
\end{minipage}
\begin{minipage}[t]{0.38\textwidth}
\includegraphics[width=\linewidth]{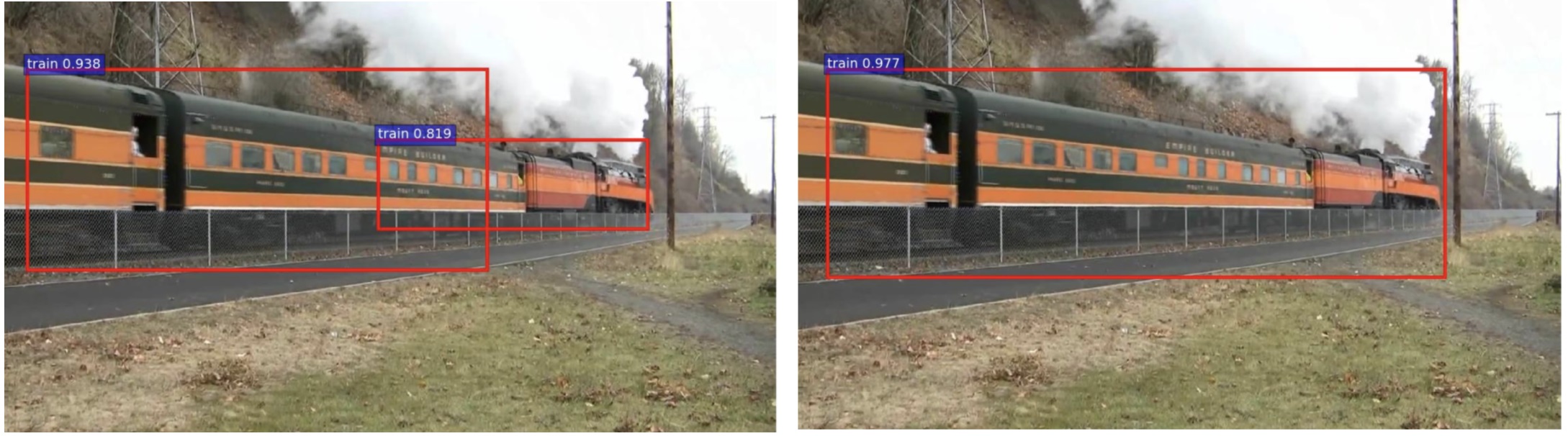}
\end{minipage}
\hspace{2mm}
\begin{minipage}[t]{0.38\textwidth}
\includegraphics[width=\linewidth]{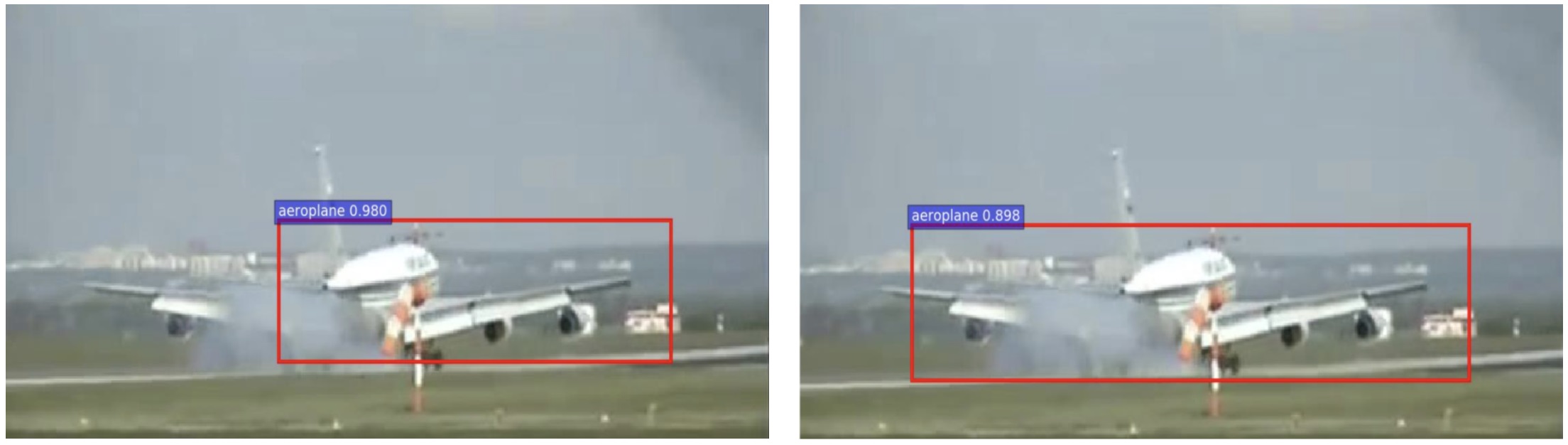}
\end{minipage}
\begin{minipage}[t]{0.38\textwidth}
\includegraphics[width=\linewidth]{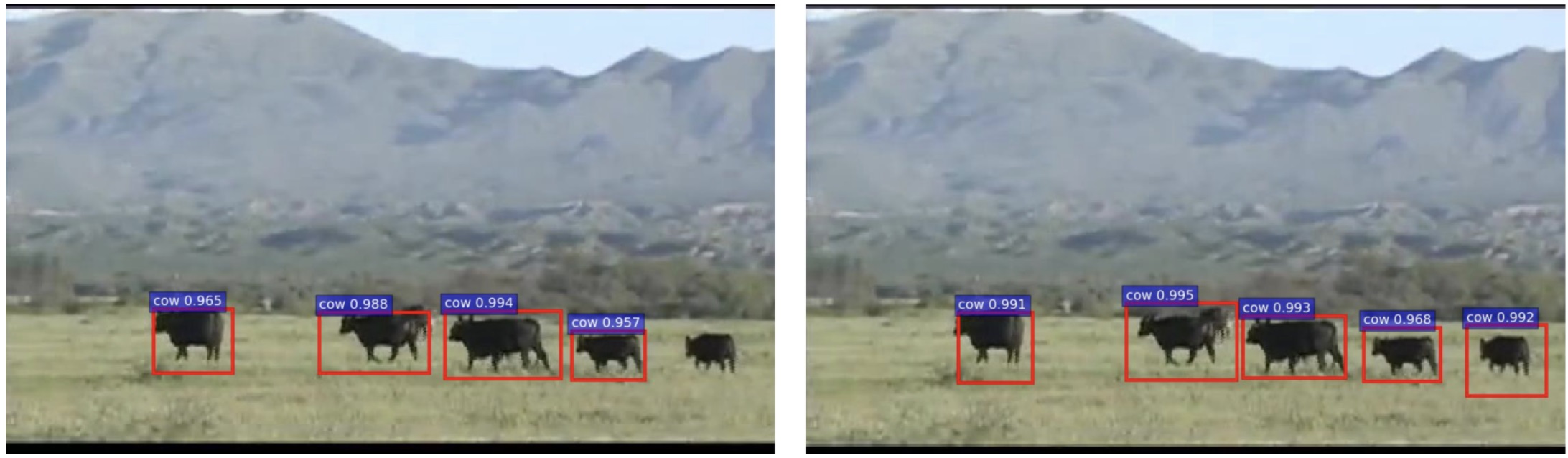}
\end{minipage}
\hspace{2mm}
\begin{minipage}[t]{0.38\textwidth}
\includegraphics[width=\linewidth]{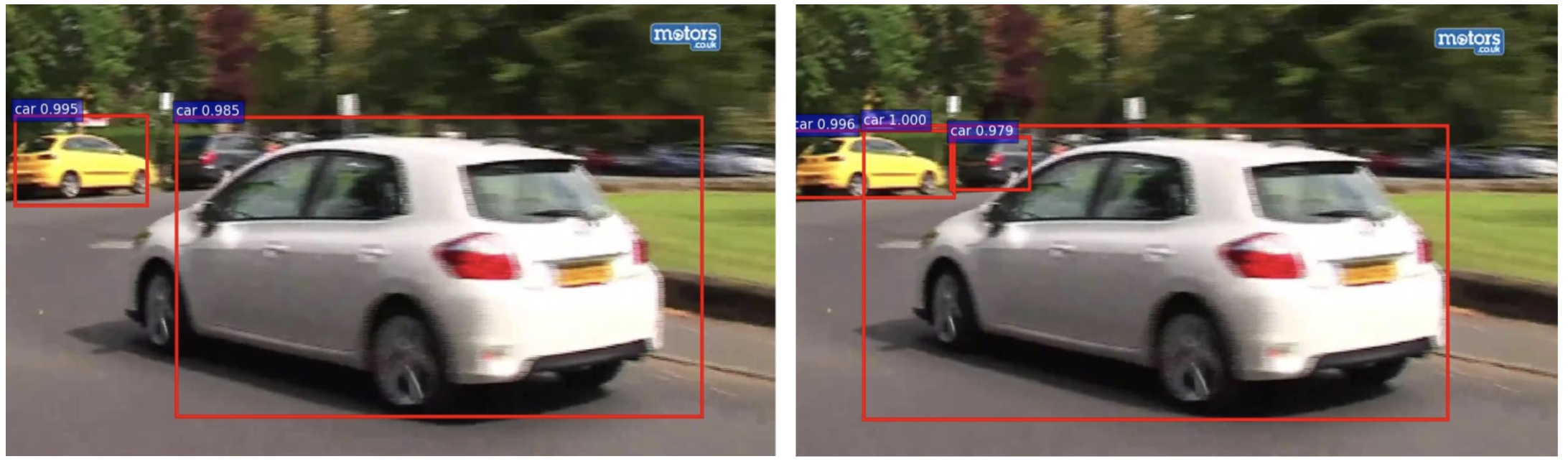}
\end{minipage}
\caption{Exemplary detections on Youtube-Objects dataset using Faster R-CNN trained on CycleGAN transformed VOC(propsed) and original VOC(baseline). Training on CycleGAN transformed dataset enables an object detector to deal with visual characteristics of videos such as cluttered background, motion blur effect, small size of objects etc., and thus improves detection performance on test set sampled from video sequences. Please note how small objects such as distant persons, bikes, animals are detected by our model. Also note how our model is robust to distant blurry objects, cluttered/occluded group of objects. All these are achieved with zero video supervision.}
\end{figure*} 
%=============
\subsection{Datasets}
\par We use fully annotated dataset of static images as source domain and a dataset of unannotated video frames as target domain. Please note in this paper we are performing object detection on stand-alone video frames and not on video sequences. Exploiting temporal information for efficient object detection is a separate genre of research and is not the main interest of the paper. 
\par For source domain we use the images from PASCAL VOC 2007 dataset \cite{everingham2010pascal}, which is a standard dataset for object detection. PASCAL VOC 2007 dataset consists of about 5000 training and 5000 testing images for 20 object categories. We train the object detector on all 20 object categories. For the target domain we use video frames from YouTube-Objects dataset \cite{prest2012learning} which consists of about 4300 training and 1800 testing images over 10 object categories which forms a subset of PASCAL VOC dataset. Testing is done on the annotated test set of Youtube-Objects dataset (YTO). For testing, we also consider another dataset, Youtube-Objects-Subset\cite{tang2013discriminative} which is derived from the videos of Youtube-Objects dataset  but has more ground truth annotations. Please note we never make use of bounding box annotations on video training set in any stage of our framework.
%====================
\subsection{CycleGAN Training}
\subsubsection{Effect of adversarial image transformation }
After completion of CycleGAN training, we would expect a high quality static image to be transformed to visually look like a video frame. In general, after the transformation, static images loose high frequency components, colors tend to get desaturated, discriminative parts gets blended with surroundings. We show some exemplary transformations in Fig. \ref{fig_effect}.
%========================
%====================
%==============================================
\begin{figure*}[!t]
    \centering
    \includegraphics[scale=0.4]{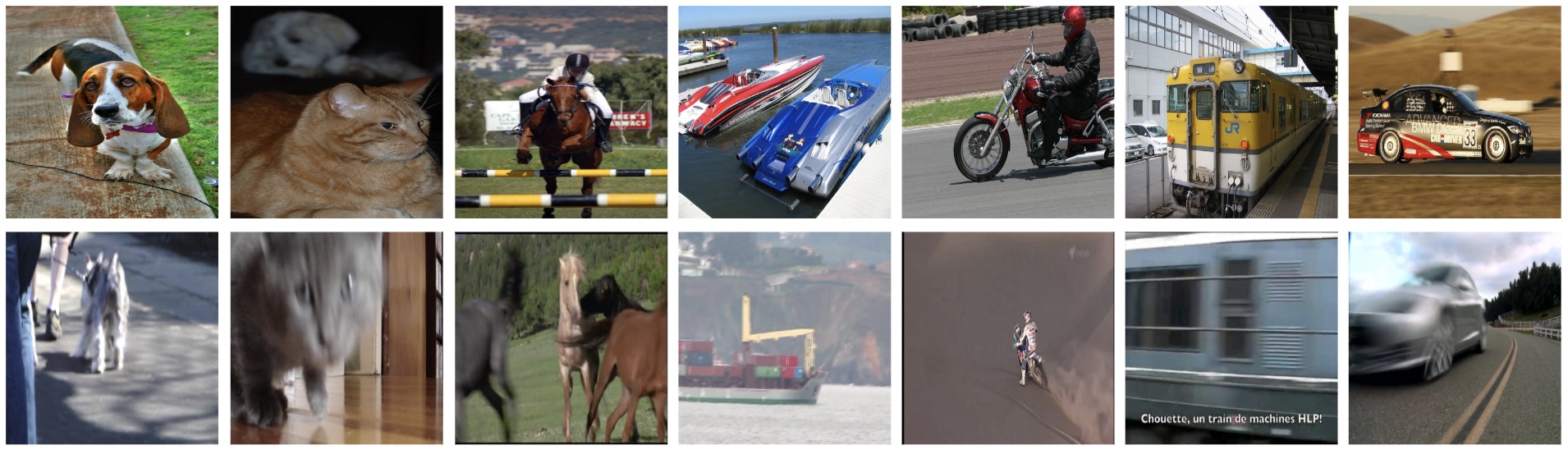}  
    \caption{Visualizing domain differences between images from VOC(top row) and YTO(bottom row) datasets. Images in VOC are sharper, rich in color representation and properly focused. Objects in YTO are usually not focused, blurred and manifests color desaturation.}%
    \label{fig_visual_difference}%
\end{figure*}
%==================================================
\subsubsection{ForwardGAN}
To appreciate the benefit of such a cyclic structure in our adversarial network, we also trained a simple forward transform GAN with only the video domain discriminator. We term this transformation as ForwardGAN. This is similar to the methods of \cite{shrivastava2017learning,bousmalis2017unsupervised}. The discriminator discriminates between video frames and forward transformed static images. As we can see in Fig. \ref{fig_gans}, ForwardGAN adds implausible structural perturbations to the images. Hence the results of training object detectors on these images gave poor test results.
%==================================
\begin{table*}[!t]
\scriptsize
  \begin{center}
  \caption { Comparison of detection performances(mean Average Precision: mAP) of Faster R-CNN object detector (trained on different variants of training datasets) on Youtube-Objects dataset }
    \begin{tabular}{c|cccccccccc|c}
      \hline
       \hspace{0.5in} Train Set \hspace{0.5in} & aero & bird & boat & car & cat & cow & dog & horse & bike & train & mAP\\
      \hline
       Original VOC(Lower Bound) & 75.0 & 89.9 & 35.1 & 68.9 & 56.7 & 46.2 & 45.7 & 39.8 & 62.6 & 46.2 & 56.6 \\
       ForwardGAN VOC & 75.3 & 83.1 & 31.4 & 68.3 & 51.0 & 55.8 & 37.6 & 41.9 & 61.5 & 47.3 & 55.3 \\
       CycleGAN VOC(\textbf{Proposed})& 77.8 & 89.6  & 34.9 & 70.9 & 58.3 & 65.6 & 45.7 & 44.7 & 62.6 & 51.4 & 60.2 \\ 
       Conditioned CycleGAN VOC(\textbf{Proposed})& 79.6 & 90.4  & 37.5 & 71.9 & 61.1 & 67.4 & 47.7 & 46.3 & 64.8 & 52.8 & 62.0 \\ 
       Youtube-Object(Upper Bound) & 78.1 & 91.3  &46.7&72.3 & 62.1 & 69.6 & 46.6 & 48.0 & 63.1 & 53.7 & 63.1\\
       \hline
       \label{tab:table-one}
    \end{tabular}
    \end{center}
\end{table*}
%=================================
%=================================
\begin{table*}[!h]
\scriptsize
  \begin{center}
  \caption {  Comparison of detection performances(mean Average Precision: mAP) of RFBNet object detector (trained on different variants of training datasets) on Youtube-Objects-Subset dataset} %
    \begin{tabular}{c|cccccccccc|c}
        \hline
        \hspace{0.5in}Train Set \hspace{0.5in} & aero & bird & boat & car & cat & cow & dog & horse & bike & train & mAP \\
        \hline
        Original VOC(Lower Bound) & 75.2 & 80.9 & 45.1 & 74.5 & 45.5 & 40.1 & 33.7 & 37.4 & 61.8 & 58.8 & 55.3 \\
        ForwardGAN VOC & 72.7 & 67.8 & 41.1 & 67.7 & 38.0 & 58.8 & 35.7 & 34.5 & 59.6 & 64.7 & 54.0 \\
        CycleGAN VOC(\textbf{Proposed}) & 82.0 & 82.6 & 48.4 & 74.7 & 49.2 & 56.3 & 34.1 & 42.0 & 60.4 & 71.0 & 60.1 \\
        Conditioned CycleGAN VOC(\textbf{Proposed}) & 83.0 & 83.8 & 49.2 & 76.1 & 51.0 & 56.7 & 35.2 & 42.7 & 61.9 & 72.1 & 61.2 \\
        Youtube-Object(Upper Bound) & 83.5 & 84.6  & 48.9 & 75.2 & 51.6 & 58.0 & 36.9	 & 45.3 & 62.7 & 74.1 & 62.1 \\ 
        \hline
        \label{tab:table-two} 
    \end{tabular}
    \end{center}
\end{table*}
%=============================
%=============================
\begin{table*}[!h]
\scriptsize
\centering
\caption {Comparison of CorLoc on Youtube-Objects and Youtube-Objects-Subset dataset. Proposed methods refer to performances of Faster R-CNN models trained on CycleGAN and class conditioned CycleGAN transformed PASCAL VOC datasets. }
% Note that all the comparing algorithms are weakly supervised for videos while our proposed method is totally unsupervised on video domain.} 
\begin{tabular}{cccccc}\hline
\multicolumn{6}{c}{Youtube Objects Dataset}                               \\\hline
Proposal Only\cite{rochan2015weakly} & Proposal + Transfer\cite{rochan2015weakly}& Teh \textit{et al.}\cite{vcip} & Chanda \textit{et al.}\cite{bmvc} & \textbf{Proposed} & \textbf{Proposed(Conditioned)}\\\hline
51.5          & 55.3                & 56.7      & 61.9         & 58.8 & 61.6   \\\hline
\multicolumn{6}{c}{Youtube Objects Subset Dataset}                        \\
36.0          & 41.6                & 48.5      & 51.1         & 50.8  & 52.1 \\ \hline
\label{table_compare}
\end{tabular}
\end{table*}
\subsubsection{Dataset adaptation or regularization ?}
Fig. \ref{fig_effect} might tempt one to believe that proposed model succeeds just by adding some noise to the training data and thereby making training data more difficult. In Fig. \ref{fig_visual_difference} we show some examples from original VOC and YTO. Indeed, we can appreciate that objects in YTO suffer from defocussing, blurriness, color desaturation effects. Our framework learns this distribution difference automatically instead of relying on manual hand crafted dataset augmentation techniques. In fact, in our initial experiments, we augmented original VOC with Gaussian noise of standard deviation  0.01, 0.05, 0.1, 0.5 and 1. mAP of Faster RCNN trained on these augmented datasets and tested on YTO were 56.4, 57.1, 57.2, 50.3 and 44.1. We also tried with Gaussian blurring of VOC images with kernel sizes of 5, 9 and 13 at standard deviation of 2. mAPs on YTO were 56.4, 57.0 and 56.8. With standard deviation of 4, mAPs were 56.3, 57.2 and 55.9. Thus, it is safe to say simple dataset augmentations do not help in the current domain adaptation problem.
%==============
\subsection{Object Detection}
\subsubsection{Evaluation Metrics}
A commonly used metric for evaluating object detection is mean Average Precision(mAP)\cite{everingham2010pascal}. According to Pascal criterion, a detected bounding box, $B_p$, is considered as true positive, $TP$, if, Intersection Over Union (IoU) = $\frac{area(B_p \cap B_{gt})}{area(B_p \cup  B_{gt})} \ge 0.5$, else $B_p$ is considered as a false positive, $FP$. $B_{gt}$ is an annotated box. mAP is defined as the mean of average precision over all classes. Another popular measure is CorLoc \cite{corloc} which is given by $\frac{TP}{TP + FP}$. CorLoc can be interpreted as the proportion of detected bounding boxes which satisfy the Pascal criterion.\\
\textbf{Special care on Youtube-Objects-Subset dataset:} The annotations released for this dataset consist of pixel-wise segmentation maps instead of bounding boxes as in Youtube-Objects. We convert the segmentation maps to bounding boxes by enclosing the segmentation maps by smallest possible rectangles. We also followed the conversion strategy as proposed in \cite{vcip}; converting a detected bounding box to a segmentation map using grab-cut algorithm \cite{rother2004grabcut}\footnote{Available: \href{https://docs.opencv.org/trunk/d8/d83/tutorial\_py\_grabcut.html}{https://docs.opencv.org/trunk/d8/d83/tutorial\_py\_grabcut.html}}. Under this formulation, IoU is measured with respect to the overlap of segmentation maps. The numerical results following both paradigms are almost comparable and thus, following \cite{vcip}, we stick to the latter framework.
%==============
\subsubsection{Comparing Baselines}
\par As a baseline we train standard object detectors on PASCAL VOC dataset and test on video frames without any adaptation. This gives us the lower bound of performance. The upper bound is achieved by training detectors on annotated video frames. To show the robustness of the approach, we report the observed effect of domain adaptation using two different object detectors, viz., Faster R-CNN and RFBNet300. We have considered backbone as VGG16 for both detectors. In Table \ref{tab:table-one} we compare the performances of Faster R-CNN trained on different versions of training set  and then tested on Youtube-Objects(YTO) test set. In Table \ref{tab:table-two} we report performances with RFBNet300 network tested on Youtube-Objects-Subset. For both detectors, we see an appreciable boost in performance if trained on CycleGAN transformed images dataset compared to original VOC dataset. Of course, training on labelled YTO frames still gives the upper bound of performance but using our proposed method we reduce the performance gap appreciably. Also, it is to be noted that the training of ForwardGAN deteriorates performance even worse than training on original VOC. Tables \ref{tab:table-one} and \ref{tab:table-two} strongly bolster our hypothesis that visual domain adaptation is a viable approach for cross domain learning of object detectors with close to zero supervision in unlabeled domain.
%==================================%===============================
Next, in Table \ref{table_compare}, we also compare our model with some of the contemporary weakly supervised baselines. In \cite{rochan2015weakly} (proposal only) refers to learning of appearance model based on object proposals on weakly annotated frames while (proposal + transfer) refers to combination of appearance model from 'novel' objects (video) and transferred appearance model from 'familiar' annotated objects (static images). Method of  Chanda \textit{et al.}\cite{bmvc} is based on a 2-stream network, wherein one stream they perform regular fully supervised image object detection and in another stream, they perform frame level classification on the weakly annotated videos. These 2 streams share parameters to counter domain shift factors. Teh \textit{et al.}\cite{vcip} proposed an attention network so as to mimic the score of region proposal networks on weakly annotated objects to be similar to a strong fully supervised object detector. Please note all these competing methods assume the presence of meta information such as presence/absence of object categories on each training video frame. However, we assume no information to be associated on video training dataset. It is evident from Table \ref{table_compare} that our model presents competitive performance(better in majority cases) compared to these methods. 
%=================================
\subsection{Can YTO class labels help CycleGAN?}
One of the drawbacks\footnote{See discussion: \href{https://github.com/junyanz/CycleGAN\#failure-cases}{https://github.com/junyanz/CycleGAN\#failure-cases}} of CycleGAN is that it is tough to train if the two domains differ structurally. In our case, this bottleneck arises because, without a priori knowledge of labels, a given mini batch can consist of different object categories. To mitigate this drawback one can train a separate CycleGAN conditioned on each category on YTO. This requires weak label information for each frame. So, during offline CycleGAN training, we train 10 different networks to individually transform each category. However, we still need to train only a single object detector on this conglomerated transformed dataset. Such use of weak labels further boosts the performance of our framework as reported in Tables \ref{tab:table-one}, \ref{tab:table-two} and \ref{table_compare}. This observation indicates that class conditioned CycleGANs are better at capturing the appearance diversities across two datasets.
\section{Conclusion}
In this paper, we mainly focused on unsupervised pixel level domain adaptation for transferring image object detectors for videos. In contrast to the contemporary trend of weakly supervised learning on videos which still requires manual intervention in videos, our framework requires no supervision. The core idea is to pose the problem as an adversarial image-to-image translation for converting annotated static images to be visually indistinguishable from video frames. We also showed that the inclusion of class labels on videos improves our framework further. A straightforward application of our model will be to automatically annotate large video datasets for object detection. Currently, our method is focused on detecting objects on standalone video frames. An immediate extension would be to leverage temporal information in videos for enhanced detection performance. Moreover, the ideas of the paper in general, should encourage researchers towards other interesting visual domain adaptation applications such as emotion recognition from 3D face avatars, learning pose estimation in a virtual world, robotic navigation in simulated environments and finally applying in real world frameworks.  
%=========================================
{\small
\bibliographystyle{ieee}
\bibliography{egbib}

\begin{thebibliography}{10}\itemsep=-1pt

\bibitem{bilen2014weakly}
H.~Bilen, M.~Pedersoli, and T.~Tuytelaars.
\newblock Weakly supervised object detection with posterior regularization.
\newblock In {\em Proceedings BMVC 2014}, pages 1--12, 2014.

\bibitem{bousmalis2017unsupervised}
K.~Bousmalis, N.~Silberman, D.~Dohan, D.~Erhan, and D.~Krishnan.
\newblock Unsupervised pixel-level domain adaptation with generative
  adversarial networks.
\newblock In {\em CVPR}, volume~1, page~7, 2017.

\bibitem{bmvc}
O.~Chanda, E.~W. Teh, M.~Rochan, Z.~Guo, and Y.~Wang.
\newblock Adapting object detectors from images to weakly labeled videos.
\newblock In {\em The 28th British Machine Vision Conference (BMVC)}.

\bibitem{dalal2005histograms}
N.~Dalal and B.~Triggs.
\newblock Histograms of oriented gradients for human detection.
\newblock In {\em Computer Vision and Pattern Recognition, 2005. CVPR 2005.
  IEEE Computer Society Conference on}, volume~1, pages 886--893. IEEE, 2005.

\bibitem{deng2009imagenet}
J.~Deng, W.~Dong, R.~Socher, L.-J. Li, K.~Li, and L.~Fei-Fei.
\newblock Imagenet: A large-scale hierarchical image database.
\newblock In {\em Computer Vision and Pattern Recognition, 2009. CVPR 2009.
  IEEE Conference on}, pages 248--255. Ieee, 2009.

\bibitem{corloc}
T.~Deselaers, B.~Alexe, and V.~Ferrari.
\newblock Weakly supervised localization and learning with generic knowledge.
\newblock {\em International journal of computer vision}, 100(3):275--293,
  2012.

\bibitem{everingham2010pascal}
M.~Everingham, L.~Van~Gool, C.~K. Williams, J.~Winn, and A.~Zisserman.
\newblock The pascal visual object classes (voc) challenge.
\newblock {\em International journal of computer vision}, 88(2):303--338, 2010.

\bibitem{felzenszwalb2010object}
P.~F. Felzenszwalb, R.~B. Girshick, D.~McAllester, and D.~Ramanan.
\newblock Object detection with discriminatively trained part-based models.
\newblock {\em IEEE transactions on pattern analysis and machine intelligence},
  32(9):1627--1645, 2010.

\bibitem{faster}
R.~Girshick.
\newblock Fast r-cnn.
\newblock In {\em Proceedings of the IEEE international conference on computer
  vision}, pages 1440--1448, 2015.

\bibitem{girshick2014rich}
R.~Girshick, J.~Donahue, T.~Darrell, and J.~Malik.
\newblock Rich feature hierarchies for accurate object detection and semantic
  segmentation.
\newblock In {\em Proceedings of the IEEE conference on computer vision and
  pattern recognition}, pages 580--587, 2014.

\bibitem{gokberk2014multi}
R.~Gokberk~Cinbis, J.~Verbeek, and C.~Schmid.
\newblock Multi-fold mil training for weakly supervised object localization.
\newblock In {\em Proceedings of the IEEE conference on computer vision and
  pattern recognition}, pages 2409--2416, 2014.

\bibitem{goodfellow2014generative}
I.~Goodfellow, J.~Pouget-Abadie, M.~Mirza, B.~Xu, D.~Warde-Farley, S.~Ozair,
  A.~Courville, and Y.~Bengio.
\newblock Generative adversarial nets.
\newblock In {\em Advances in neural information processing systems}, pages
  2672--2680, 2014.

\bibitem{he2016deep}
K.~He, X.~Zhang, S.~Ren, and J.~Sun.
\newblock Deep residual learning for image recognition.
\newblock In {\em Proceedings of the IEEE conference on computer vision and
  pattern recognition}, pages 770--778, 2016.

\bibitem{pix2pix2017}
P.~Isola, J.-Y. Zhu, T.~Zhou, and A.~A. Efros.
\newblock Image-to-image translation with conditional adversarial networks.
\newblock {\em CVPR}, 2017.

\bibitem{kalogeiton2016analysing}
V.~Kalogeiton, V.~Ferrari, and C.~Schmid.
\newblock Analysing domain shift factors between videos and images for object
  detection.
\newblock {\em IEEE Trans. Pattern Anal. Mach. Intell.}, 38(11):2327--2334,
  2016.

\bibitem{kingma2014adam}
D.~P. Kingma and J.~Ba.
\newblock Adam: A method for stochastic optimization.
\newblock {\em arXiv preprint arXiv:1412.6980}, 2014.

\bibitem{kumar2010self}
M.~P. Kumar, B.~Packer, and D.~Koller.
\newblock Self-paced learning for latent variable models.
\newblock In {\em Advances in Neural Information Processing Systems}, pages
  1189--1197, 2010.

\bibitem{leistner2011improving}
C.~Leistner, M.~Godec, S.~Schulter, A.~Saffari, M.~Werlberger, and H.~Bischof.
\newblock Improving classifiers with unlabeled weakly-related videos.
\newblock 2011.

\bibitem{li2016precomputed}
C.~Li and M.~Wand.
\newblock Precomputed real-time texture synthesis with markovian generative
  adversarial networks.
\newblock In {\em European Conference on Computer Vision}, pages 702--716.
  Springer, 2016.

\bibitem{lin2014microsoft}
T.-Y. Lin, M.~Maire, S.~Belongie, J.~Hays, P.~Perona, D.~Ramanan,
  P.~Doll{\'a}r, and C.~L. Zitnick.
\newblock Microsoft coco: Common objects in context.
\newblock In {\em European conference on computer vision}, pages 740--755.
  Springer, 2014.

\bibitem{liu2017receptive}
S.~Liu, D.~Huang, and Y.~Wang.
\newblock Receptive field block net for accurate and fast object detection.
\newblock {\em arXiv preprint arXiv:1711.07767}, 2017.

\bibitem{malisiewicz2011ensemble}
T.~Malisiewicz, A.~Gupta, and A.~A. Efros.
\newblock Ensemble of exemplar-svms for object detection and beyond.
\newblock In {\em Computer Vision (ICCV), 2011 IEEE International Conference
  on}, pages 89--96. IEEE, 2011.

\bibitem{prest2012learning}
A.~Prest, C.~Leistner, J.~Civera, C.~Schmid, and V.~Ferrari.
\newblock Learning object class detectors from weakly annotated video.
\newblock In {\em Computer Vision and Pattern Recognition (CVPR), 2012 IEEE
  Conference on}, pages 3282--3289. IEEE, 2012.

\bibitem{rochan2015weakly}
M.~Rochan and Y.~Wang.
\newblock Weakly supervised localization of novel objects using appearance
  transfer.
\newblock In {\em Proceedings of the IEEE Conference on Computer Vision and
  Pattern Recognition}, pages 4315--4324, 2015.

\bibitem{rother2004grabcut}
C.~Rother, V.~Kolmogorov, and A.~Blake.
\newblock Grabcut: Interactive foreground extraction using iterated graph cuts.
\newblock In {\em ACM transactions on graphics (TOG)}, volume~23, pages
  309--314. ACM, 2004.

\bibitem{sharma2013efficient}
P.~Sharma and R.~Nevatia.
\newblock Efficient detector adaptation for object detection in a video.
\newblock In {\em Proceedings of the IEEE Conference on Computer Vision and
  Pattern Recognition}, pages 3254--3261, 2013.

\bibitem{shrivastava2017learning}
A.~Shrivastava, T.~Pfister, O.~Tuzel, J.~Susskind, W.~Wang, and R.~Webb.
\newblock Learning from simulated and unsupervised images through adversarial
  training.
\newblock In {\em The IEEE Conference on Computer Vision and Pattern
  Recognition (CVPR)}, volume~3, page~6, 2017.

\bibitem{song2014learning}
H.~O. Song, R.~Girshick, S.~Jegelka, J.~Mairal, Z.~Harchaoui, and T.~Darrell.
\newblock On learning to localize objects with minimal supervision.
\newblock {\em arXiv preprint arXiv:1403.1024}, 2014.

\bibitem{tang2013discriminative}
K.~Tang, R.~Sukthankar, J.~Yagnik, and L.~Fei-Fei.
\newblock Discriminative segment annotation in weakly labeled video.
\newblock In {\em Proceedings of the IEEE conference on computer vision and
  pattern recognition}, pages 2483--2490, 2013.

\bibitem{vcip}
E.~W. Teh, Z.~Guo, and Y.~Wang.
\newblock Object localization in weakly labeled data using regularized
  attention networks.
\newblock In {\em Visual Communications and Image Processing (VCIP), 2017
  IEEE}, pages 1--4. IEEE, 2017.

\bibitem{wohlhart2015learning}
P.~Wohlhart and V.~Lepetit.
\newblock Learning descriptors for object recognition and 3d pose estimation.
\newblock In {\em CVPR}, pages 3109--3118, 2015.

\bibitem{zhu2017unpaired}
J.-Y. Zhu, T.~Park, P.~Isola, and A.~A. Efros.
\newblock Unpaired image-to-image translation using cycle-consistent
  adversarial networks.
\newblock {\em arXiv preprint arXiv:1703.10593}, 2017.

\end{thebibliography}
}

\end{document}